\documentclass[runningheads]{llncs}
\usepackage[table, dvipsnames]{xcolor}
 
\usepackage{eccv}

\usepackage{eccvabbrv}

\usepackage{graphicx}
\usepackage{booktabs}

\usepackage[accsupp]{axessibility}  

\usepackage{multirow}
\usepackage{wrapfig}

\usepackage{hyperref}

\usepackage{orcidlink}

\newcommand{\nbf}[1]{\noindent \textbf{#1.}}

\definecolor{Gray}{gray}{0.85}

\usepackage{algorithm}
\usepackage{algorithmic}

\usepackage{pifont}

\newcommand{\nohave}{\color{red} \ding{55}}
\newcommand{\have}{\color{green} \ding{51}}

\makeatletter
  \newcommand\figcaption{\def\@captype{figure}\caption}
  \newcommand\tabcaption{\def\@captype{table}\caption}
\makeatother

\begin{document}

\title{Dual-stage Hyperspectral Image Classification Model with Spectral Supertoken} 

\titlerunning{DSTC: Dual-stage Spectral Supertoken Classifier}

\makeatletter
\def\thanks#1{\protected@xdef\@thanks{\@thanks
        \protect\footnotetext{#1}}}
\makeatother

\author{Peifu Liu\orcidlink{0000-0003-4018-1478} \and
Tingfa Xu$^{\dag}$\orcidlink{0000-0001-5452-2662} \and
Jie Wang\orcidlink{0000-0002-4847-3697} \and
Huan Chen\orcidlink{0000-0003-1965-9107} \and
Huiyan Bai\orcidlink{0009-0000-8037-2338} \and 
Jianan Li$^{\dag}$\orcidlink{0000-0002-6936-9485} 
\thanks{$^{\dag}$ Correspondence to: Tingfa Xu and Jianan Li.} 
}

\authorrunning{P. Liu et al.}

\institute{Beijing Institute of Technology}

\maketitle

\begin{abstract}
Hyperspectral image classification, a task that assigns pre-defined classes to each pixel in a hyperspectral image of remote sensing scenes, often faces challenges due to the neglect of correlations between spectrally similar pixels. This oversight can lead to inaccurate edge definitions and difficulties in managing minor spectral variations in contiguous areas. To address these issues, we introduce the novel Dual-stage Spectral Supertoken Classifier (DSTC), inspired by superpixel concepts. DSTC employs spectrum-derivative-based pixel clustering to group pixels with similar spectral characteristics into spectral supertokens. By projecting the classification of these tokens onto the image space, we achieve pixel-level results that maintain regional classification consistency and precise boundary. Moreover, recognizing the diversity within tokens, we propose a class-proportion-based soft label. This label adaptively assigns weights to different categories based on their prevalence, effectively managing data distribution imbalances and enhancing classification performance. Comprehensive experiments on WHU-OHS, IP, KSC, and UP datasets corroborate the robust classification capabilities of DSTC and the effectiveness of its individual components. Code will be publicly available at \url{https://github.com/laprf/DSTC}.

\keywords{Dual-stage Spectral Supertoken Classifier \and Hyperspectral Image Classification \and Clustering}
\end{abstract}

\section{Introduction}
\label{sec:intro}

\begin{figure}[htp]
    \centering
    \includegraphics[width=0.9\linewidth]{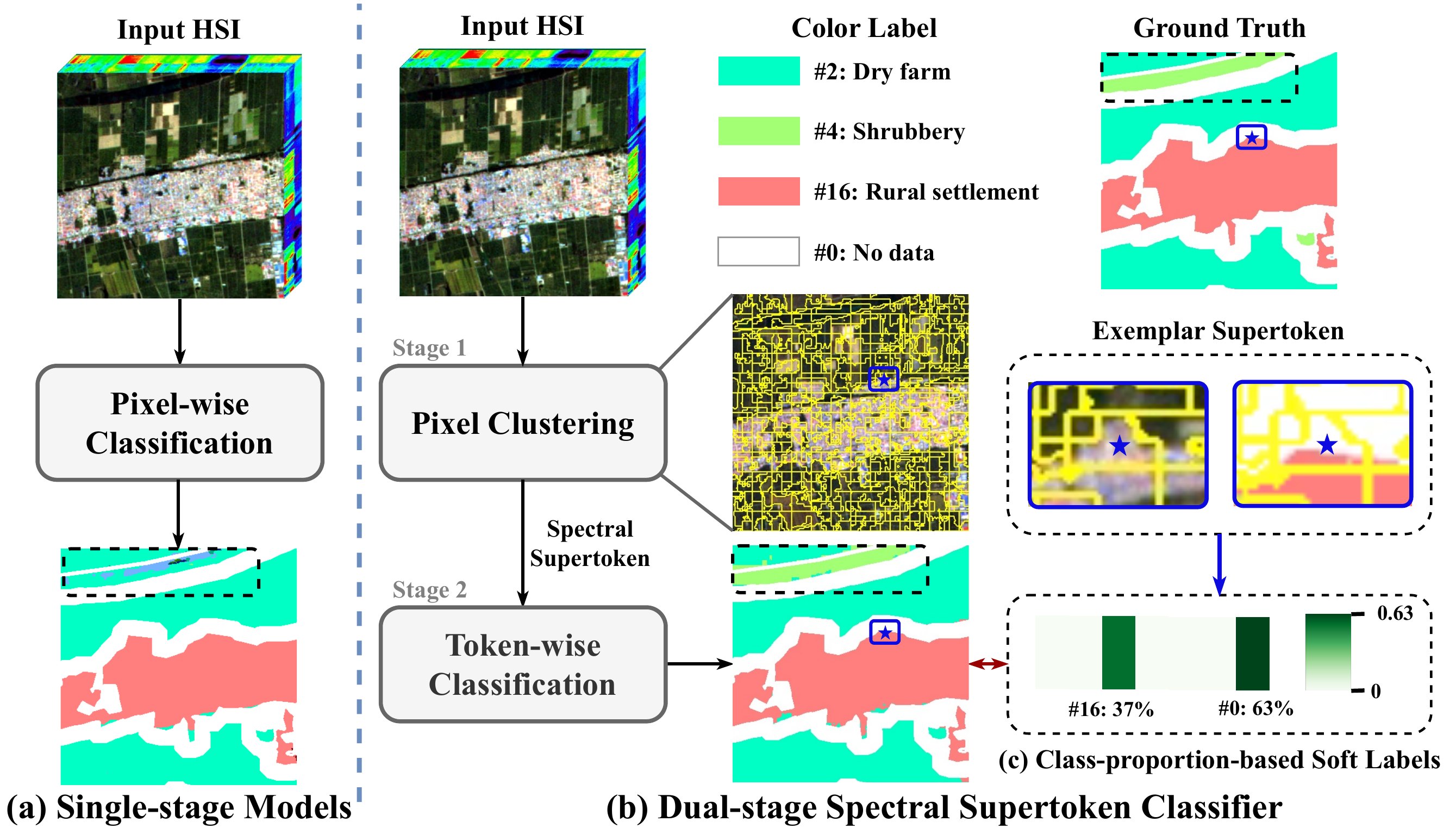}
    \caption{\textbf{(a)} Single-stage pixel-wise classification models exhibit limitations in handling minor spectral variations and fail to deliver precise boundary delineation, as showcased in the black-framed area. In contrast, our \textbf{(b)} Dual-stage Spectral Supertoken Classifier effectively clusters similar pixels into spectral supertokens for token-wise classification, yielding improved classification within contiguous regions. This process is enhanced by \textbf{(c)} class-proportion-based soft labels, extracted from the proportion of each land cover type within each supertoken boundaries (illustrating with blue supertoken's example).
   }
    \label{fig: motivation}
\end{figure}

Hyperspectral imaging concurrently captures spatial and spectral information of a target scene, thereby producing a hyperspectral image (HSI) that embodies a robust fusion of spatial-spectral data. Within the domain of remote sensing, hyperspectral image classification (HSIC) is crucial for assigning pre-defined land cover categories to each pixel, based on its spectral response and spatial characteristics~\cite{Quantum}. This classification process is integral to various sectors such as urban construction~\cite{urban}, agriculture~\cite{Agriculture}, and mining~\cite{mining}, \etc.

Deep learning methods for hyperspectral image classification, specifically encoder-decoder models, are recognized for their feature extraction capabilities. However, these techniques frequently overlook the spectral similarities between adjacent pixels, favoring individual pixel-level classification instead. This approach gives rise to two primary issues. \textbf{(i)} It results in blurred or incorrect demarcations between regions with different categories. \textbf{(ii)} It yields inconsistent classification for pixels within contiguous areas, even when these areas exhibit only minor spectral variations, as depicted in \cref{fig: motivation}~(a).

Numerous studies have endeavored to tackle these challenges by employing superpixel algorithms. Zhao~\etal~\cite{Superpixel-Guided} utilized the SLIC~\cite{SLIC} algorithm to generate superpixel segmentation, subsequently reshape convolutional kernels to enhance edge learning. However, this approach fails to ensure the consistency of the classification results within contiguous regions. Additionally, the iterative nature of the SLIC algorithm, coupled with its incompatibility with CUDA acceleration, limits its suitability for real-time inference.

To preserve accurate contours, ensure consistent region classification, and maintain efficient computation, we draw inspiration from Faster R-CNN~\cite{Faster_RCNN}. We propose a novel two-stage model, named the Dual-stage Spectral Supertoken Classifier (DSTC), which is designed for real-time and accurate hyperspectral image classification. As illustrated in \cref{fig: motivation}~(b), stage 1 involves spectrally similar pixel clustering, which generates spectral supertokens (SSTs). These supertokens are subsequently categorized into pre-defined classes in stage 2.

More specifically, in stage 1, our method utilizes spectrum-derivative-based pixel clustering. This process involves the computation of spectral derivative features that enrich the original spectra. Subsequently, we assess the spectral similarity of each pixel in relation to the clustering centers, which aids in grouping pixels into clusters. This pre-classification strategy considerably reduces the probability of error and ensures more precise and distinct boundary definitions. Following this, we consolidate the semantic features within each cluster to form spectral supertokens. Since each token represents a multitude of pixels, this approach significantly diminishes the number of image primitives.

Given that a single supertoken may encompass various ground objects, a simplistic method would be to use the average response of the dominant type within such a group as the training label. However, this could overlook less prevalent categories. To overcome this limitation, we introduce a class-proportion-based soft label (CPSL). Instead of assigning a definitive class to each token, CPSL discloses the diverse land covers present within each spectral supertoken and their respective proportions, as depicted in \cref{fig: motivation}~(c). This labeling approach effectively mitigates the challenges associated with the uneven distribution of land cover types, thereby significantly improving the model's classification accuracy.

In conclusion, our DSTC initiates by aggregating local features and subsequently investigating inter-token long-range relationships. This two-phase model amalgamates local and global insights, delivering robust classification performance with minimal computational expense.

We evaluate the effectiveness of our DSTC by testing it on the WHU-OHS dataset~\cite{WHU-OHS}, as well as on the IP, KSC, and UP datasets. The results indicate DSTC's robust classification performance on larger scale and higher resolution datasets. Additionally, we conduct an extended experiment on the salient object detection dataset HS-SOD~\cite{HS-SOD}, showcasing DSTC's generalization capability.

In a nutshell, our contributions can be summarized as follows:
\begin{itemize}
    \item[$\bullet$] We introduce a novel Dual-stage Spectral Supertoken Classifier. To the best of our knowledge, this is the first attempt to develop a two-stage, end-to-end trainable deep neural network for hyperspectral image classification.
    
    \item[$\bullet$] We incorporate a spectrum-derivative-based pixel clustering technique, which pre-classifies spectrally similar pixels, with the aim of enhancing classification accuracy and reducing computational cost.
    
    \item[$\bullet$] We propose a class-proportion-based soft label, a uniquely designed supervision method, to counteract the adverse effects of uneven data distribution.
\end{itemize}

\section{Related Work}
\nbf{Hyperspectral Image Classification}
Traditional machine learning methods utilize statistical learning to analyze data distribution and develop classification models~\cite{melgani2004classification, pesaresi2008robust, li2013generalized}. Deep learning has made considerable strides, especially in extracting spatial-spectral features~\cite{1D-CNN, 3D-CNN, FreeNet, CLSJE}. However, despite these advancements, existing models tend to overlook pixel similarity, leading to blurred edges and difficulties in managing minor spectral changes. Two-stage methodologies may offer a solution to these challenges. Tu~\etal~\cite{tu2019dual} first proposed the dual-stage construction of probability for hyperspectral image classification. They extracted the shape attributes of HSI to establish the initial classification probability map, which was subsequently refined using a rolling guidance filter. Despite effectively preserving boundaries, this approach showed suboptimal classification performance and high computational cost due to its reliance on hand-crafted features. To address these limitations, we propose DSTC, the first two-stage, end-to-end trainable deep model for HSI classification. DSTC efficiently clusters spectrally similar pixels into supertokens and performs token-wise classification, overcoming the drawbacks of previous methods.

\nbf{Hyperspectral Superpixel Clustering}
Clustering algorithms can be broadly classified into traditional and deep learning-based categories. Traditional methods, such as SLIC~\cite{SLIC} and manifold-SLIC~\cite{manifold-slic}, depend on handcrafted features and often exhibit limited performance. The advent of deep learning has facilitated the development of advanced clustering techniques~\cite{cluster1, cluster2, Super_sampling_network, SVit}, such as SSN~\cite{Super_sampling_network}, which computes superpixels using a differentiable model, and SViT~\cite{SVit}, which incorporates an attention mechanism for sparse sampling. In the realm of HSI analysis, Zhao~\etal~\cite{MAT-ASSAL} utilized Principal Component Analysis for dimensionality reduction, while Barbato~\etal~\cite{BARBATO2022100823} improved SLIC by amalgamating spectral and spatial data for clustering. Zhang~\etal~\cite{9718213} merged dimensionality reduction with advanced clustering to enhance feature learning. However, these methods typically necessitate numerous iterations, which impedes real-time application. In our approach, we incorporate a novel pixel clustering algorithm into our model to expedite computation through the use of CUDA acceleration.

\nbf{Superpixel in Hyperspectral Image Classification}
Local feature descriptors are extracted and clustered using clustering algorithms. Each cluster center represents a "visual vocabulary" that can be applied to downstream tasks~\cite{8887285}. These early applications inspired the use of superpixels in hyperspectral image classification.
Machine learning methods combine superpixels with morphological filtering~\cite{6100425}, support vector machines~\cite{6297992, 6723581, 6860780}, and Markov random fields~\cite{6723581}, effectively mitigating the Hughes phenomenon induced by inadequate classifier performance~\cite{9328197}. In recent deep learning approaches, superpixels have also been extensively leveraged. For example, Zhao~\etal~\cite{9782104} used superpixels to guide deformable convolution kernels for feature extraction, while Tu~\etal~\cite{10115248} and Nartey~\etal~\cite{10273416} integrated superpixels with graph neural networks to enhance boundary learning. However, these methods underutilize spectral information and overlook semantic content. Moreover, their pixel-wise classification approach fails to maintain consistency within superpixels. Our method, in contrast, classifies spectral supertokens. Pixel-wise classification is achieved by assigning all pixels within a supertoken to its category.

\nbf{Transformer in Hyperspectral Image Classification}
Transformer, recognized for its attention mechanism's capacity to capture long-range interactions~\cite{attention}, has emerged as a leading model in hyperspectral classification tasks~\cite{WANG2022103005, 10167502, 10112639, SpectralFormer, Scheibenreif_2023_CVPR}. For example, Hong~\etal~\cite{SpectralFormer} segmented the input HSI along the spectral dimension and employed Transformer to generate classification results. Scheibenreif~\etal~\cite{Scheibenreif_2023_CVPR} used spatial-spectral patches with random masking for self-supervised pre-training, followed by fine-tuning during the classification stage. These methods primarily utilize Transformer as the model's core, demonstrating limited abilities in modeling local features. In contrast, our model initially extracts local features through clustering and subsequently employs the Transformer to produce token-wise classification results, effectively combining local information with the global modeling capability of Transformer.

\section{Method}
Let $\boldsymbol{I} \in \mathbb{R}^{\mathrm{H} \times \mathrm{W} \times \mathrm{D}}$ denote an existing HSI. Our goal is to create a mapping function $\boldsymbol{\Phi} (\cdot)$ that accepts $\boldsymbol{I}$ as input and outputs a classification result map $\boldsymbol{M} \in \mathbb{R}^{\mathrm{H} \times \mathrm{W}}$. $\mathrm{H}$, $\mathrm{W}$, and $\mathrm{D}$ represent the height, width, and number of spectral bands of the HSI, respectively. This process can be formulated as:
\begin{equation}
\boldsymbol{M} = \boldsymbol{\Phi} (\boldsymbol{I}).
\end{equation}

We introduce a novel Dual-stage Spectral Supertoken Classifier (DSTC) as a solution to implement this mapping function. \cref{fig: overall} presents its pipeline. The first stage encompasses the extraction of deep semantic information from input HSI. Subsequently, it aggregates this semantic information based on spectral similarity to formulate spectral supertokens. In the second stage, Transformer is utilized to classify these tokens, followed by projecting them back into the image space to obtain the final classification map.

\subsection{Spatial-preserved Feature Encoder}
\begin{figure}[tp]
    \centering
    \includegraphics[width=1\linewidth]{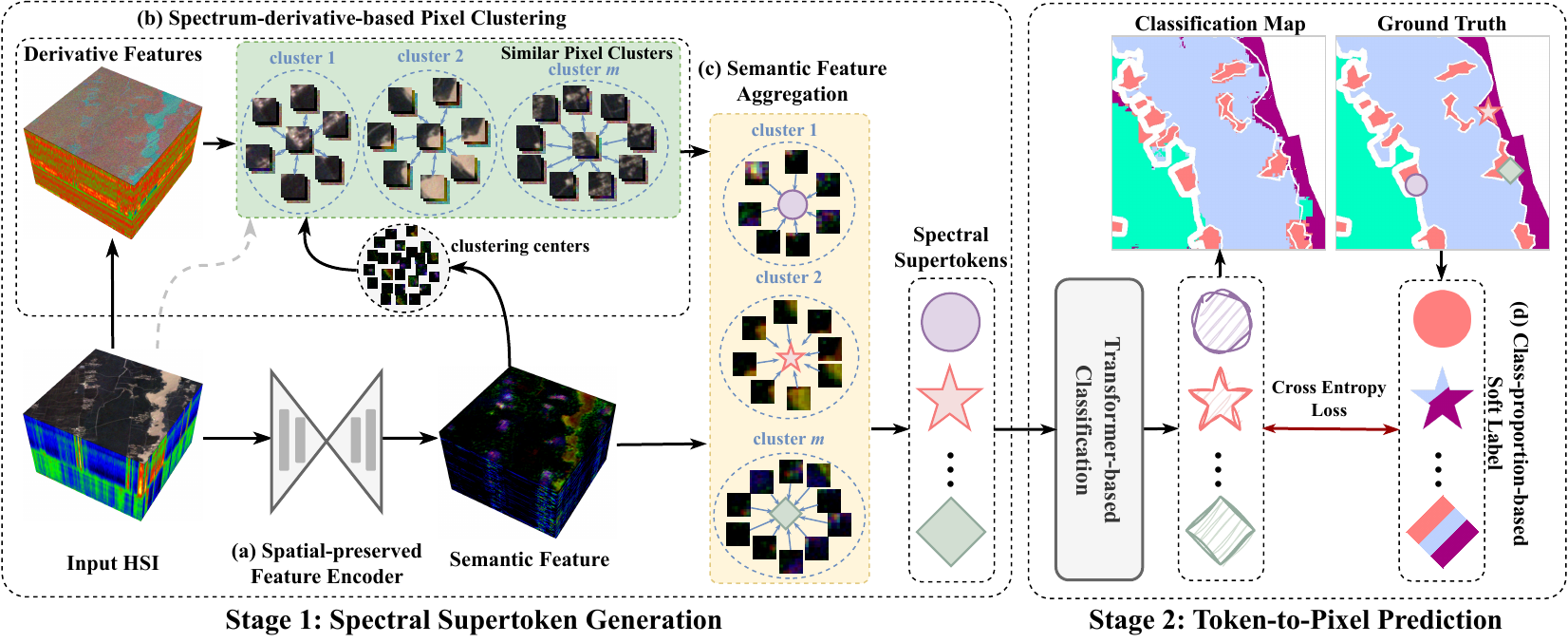}
    \caption{
    Dual-stage Spectral Supertoken Classifier begins by extracting semantic features through \textbf{(a)} spatial-preserved feature encoder. It then groups similar pixels using \textbf{(b)} spectrum-derivative-based pixel clustering. The spectral supertokens are obtained through \textbf{(c)} semantic feature aggregation. These tokens are subsequently classified by Transformer. The final classification map is generated by projecting these token-wise classifications back into the image space. Each token is supervised by a \textbf{(d)} class-proportion-based soft label during training. The varying colors in the soft labels represent different land cover types, with the proportions of these colors reflecting the respective presence of each land cover within each token.
    }
    \label{fig: overall}
\end{figure}

We commence by extracting deep semantic features \(\boldsymbol{F}_\text{D} \in \mathbb{R}^{\mathrm{H} \times \mathrm{W} \times \mathrm{C}_1}\):
\begin{equation}
    \boldsymbol{F}_\text{D} = \boldsymbol{f}_\text{se}(\boldsymbol{I}),
\end{equation}
where \(\mathrm{C}_1\) denotes the feature dimension, and \(\boldsymbol{f}_\text{se}(\cdot)\) represents the semantic extraction function. We implement this function utilizing a model based on the UNet~\cite{UNet} architecture, as illustrated in \cref{fig: overall}~(a). The model employs an off-the-shelf backbone network for downsampling, while the upsampling procedure involves a series of stacked convolutional layers. Such a design not only preserves spatial resolution but also expands the feature dimension, thereby enriching the captured semantic information.

\subsection{Spectrum-derivative-based Pixel Clustering}
This step clusters spectrally similar pixels. As illustrated in \cref{fig: overall}~(b), our initial step involves computing spectral derivative features (SDFs), which encompass a set of spectral derivatives at various orders. The first order SDF helps to separate peaks that overlap in the original spectrum, and the second order reveals intricate complex spectral details. Following this, we utilize both the original HSI and SDFs to determine the affiliation of each pixel with cluster centroids. These centroids are derived from deep semantic features.

\nbf{Spectral Derivative Features}
Consider \(\boldsymbol{W}_i \in \mathbb{R}^{\mathrm{H} \times \mathrm{W}}\) as an individual spectral band of the input HSI \(\boldsymbol{I}\). The first-order spectral derivative at the \(i\)-th band $\boldsymbol{W}^\prime_i \in \mathbb{R}^{\mathrm{H} \times \mathrm{W}}$ is computed as follows:
\begin{equation}
    \boldsymbol{W}^\prime_i = \frac{\boldsymbol{W}_j - \boldsymbol{W}_i}{j-i}=\frac{\boldsymbol{W}_j - \boldsymbol{W}_i}{\Delta n},
\end{equation}
where \(j = i + \Delta n\) and \(\Delta n > 0\) represents the step length. Similarly, the second-order spectral derivative at the \(i\)-th band $\boldsymbol{W}^{\prime \prime}_i \in \mathbb{R}^{\mathrm{H} \times \mathrm{W}}$ is obtained as:
\begin{equation}
    \boldsymbol{W}^{\prime \prime}_i = 
     \frac{\boldsymbol{W}^\prime_j - \boldsymbol{W}^\prime _i}{j-i}
     =\frac{1}{\Delta n^2} \left [ \boldsymbol{W}_{i+2\Delta n} - 2\boldsymbol{W}_{i+\Delta n} + \boldsymbol{W}_{i}\right ],
\end{equation}
where \(k = j + \Delta n = i + 2 \Delta n\). When this procedure is applied to each band, it generates the first-order spectral derivative \(\boldsymbol{I}^\prime \in \mathbb{R}^{\mathrm{H} \times \mathrm{W} \times (\mathrm{D}- \Delta n)}\), as well as the second-order spectral derivative \(\boldsymbol{I}^{\prime \prime} \in \mathbb{R}^{\mathrm{H} \times \mathrm{W} \times (\mathrm{D}- 2\Delta n)}\).

\nbf{Similar Pixel Clustering}
Our clustering methodology commences by selecting an initial set of center points. We compute the correlation of each pixel with these centroids, grouping pixels based on maximal similarity to each center, thereby forming clusters of similar pixels.

\noindent\textit{Computing Association Matrix.} 
Adopting the SLIC approach~\cite{SLIC}, we initially sample \(\mathrm{M}\) points uniformly from the deep semantic feature \(\boldsymbol{F}_\text{D}\) and compute the average features of their \(k\)-nearest neighbors, resulting in \(\mathrm{M}\) initial center points for clustering, denoted as \(\left \{ \boldsymbol{P}^0_i \right \}_{i=1}^\mathrm{M}\). Subsequently, we reshape the HSI and spectral derivative features and standardize their dimensions using a learnable linear mapping. This process yields $\boldsymbol{I}_\text{a}$, $\boldsymbol{I}^\prime_\text{a}$, and $\boldsymbol{I}^{\prime \prime}_\text{a}$, each with dimensions of $\mathrm{N} \times \mathrm{C}_2$, where $\mathrm{N} = \mathrm{H} \times \mathrm{W}$ represents the total number of image primitives, and $\mathrm{C}_2$ represents feature dimension. The association matrix \(\boldsymbol{A}^t \in \mathbb{R}^{\mathrm{N} \times \mathrm{M}}\), indicating the relationship between \(\boldsymbol{F}_\text{D}\) and the center points, is iteratively computed as:
\begin{equation}
    \label{eq: association matrix}
    \boldsymbol{A}^t(ij) = e^{-{\left \| 
    \boldsymbol{F}_\text{D}(i) + 
    \boldsymbol{I}_\text{a}(i) + 
    \boldsymbol{I}^\prime_\text{a}(i) -
    \boldsymbol{P}^{t-1}(j) 
    \right \|}^2},
\end{equation}
where \(\boldsymbol{A}^t(ij)\) is the element at the \(i\)-th row and \(j\)-th column of the association matrix in the \(t\)-th iteration. Notably, the second-order spectral derivative $\boldsymbol{I}^{\prime \prime}$ is not used, and associations are calculated only between each point and its nearby centers to enhance efficiency. 

\noindent \textit{Center Feature Update.} During iterations, the feature of each center is updated via a weighted sum of semantic feature points:
\begin{equation}
    \boldsymbol{P}^t = (\hat{\boldsymbol{A}^t})^\top \boldsymbol{F}_\text{D},
\end{equation}
with \(\hat{\boldsymbol{A}^t}\) representing the column-normalized association matrix $\boldsymbol{A}^t$. This iterative procedure, spanning \(\mathrm{T}\) iterations, culminates in the final association matrix \(\boldsymbol{A}^\mathrm{T}\) and the updated center features \(\boldsymbol{P}^\mathrm{T}\). For brevity, subsequent sections will omit the superscript notation. Finally, each point is allocated to the center with the highest association score, and each HSI or SDF is clustered into $\mathrm{M}$ groups.

\subsection{Semantic Feature Aggregation}
The concluding phase of the first stage entails the dynamic aggregation of semantic features. \cref{fig: overall}~(c) demonstrates this process. In line with Context Cluster~\cite{Context_cluster}, this process involves aggregating the semantic attributes of all points in a cluster, weighted by their similarity to the center point.

Suppose a center \(\boldsymbol{p} \in \mathbb{R}^{\mathrm{C}_2}\) is assigned \(n\) deep semantic feature points. The association vector between these \(n\) points and the center is denoted as \(\boldsymbol{a} \in \mathbb{R}^{n}\). The aggregated feature \(\boldsymbol{s} \in \mathbb{R}^{\mathrm{C}_2}\) is then computed using:
\begin{equation}
    \boldsymbol{s} = \frac{\boldsymbol{p} + \sum_{i=1}^{n} \boldsymbol{a}_i \cdot \boldsymbol{f}_{i}}{1 + \sum_{i=1}^{n} \boldsymbol{a}_i},
\end{equation}
where \(\boldsymbol{f}_{i}\) represents the \(i\)-th deep semantic feature point. This aggregation methodology is replicated for each center point, resulting in the formation of spectral supertokens $\boldsymbol{S} \in \mathbb{R}^{\mathrm{M} \times \mathrm{C}_2}$. 

\subsection{Token-to-Pixel Prediction}
We leverage Transformer's capability for feature representation and global context modeling to predict the class of each spectral supertoken:
\begin{equation}
    \hat{\boldsymbol{S}} = \boldsymbol{f}_\text{t}(\boldsymbol{S}),
\end{equation}
where \(\boldsymbol{f}_\text{t}(\cdot)\) denotes the class prediction function, implemented by the Vision Transformer (ViT)~\cite{ViT}. In this context, the shape of $\hat{\boldsymbol{S}}$ is $\mathrm{M} \times \mathrm{C}^\prime$, where $\mathrm{C}^\prime$ represents the number of pre-defined classes.

In our approach, we utilize spectral supertokens as inputs, thereby bypassing the original patch partitioning step in ViT. This is followed by employing standard attention mechanisms to elucidate the relationships between tokens. For example, in the first attention block, self-attention computation is executed as:
\begin{equation}
    \boldsymbol{S}^\prime = \boldsymbol{\sigma}(\frac{\boldsymbol{Q}_\text{S} \boldsymbol{K}^\top_\text{S}}{\sqrt{\mathrm{C}}}) \boldsymbol{V}_\text{S},
\end{equation}
where $\boldsymbol{\sigma}(\cdot)$ is the Softmax activation function. The \textit{query} $\boldsymbol{Q}_\text{S}$, \textit{key} $\boldsymbol{K}_\text{S}$, and \textit{value} $\boldsymbol{V}_\text{S}$ are obtained as follows:
\begin{equation}
    \boldsymbol{Q}_\text{S} = \boldsymbol{S} \boldsymbol{E}_\text{Q}, 
    \boldsymbol{K}_\text{S} = \boldsymbol{S} \boldsymbol{E}_\text{K}, 
    \boldsymbol{V}_\text{S} = \boldsymbol{S} \boldsymbol{E}_\text{V}.
\end{equation}
Here, $\boldsymbol{E}_\text{Q}$, $\boldsymbol{E}_\text{K}$, and $\boldsymbol{E}_\text{V}$ are the learnable parameters implemented by linear projection layers. Multi-head configurations are omitted here for brevity.

Finally, after processing through the stacked attention blocks, token-wise results $\hat{\boldsymbol{S}}$ are obtained via a linear projection. These are then transformed back to the image domain, resulting in a pixel-level classification map $\boldsymbol{M}$.

\subsection{Class-proportion-based Soft Label}
Considering that a single token may encompass multiple ground object categories, we propose a class-proportion-based soft label for supervision, as depicted in \cref{fig: overall}~(d). These soft labels are extracted from the proportion of each land cover type within supertoken boundaries. Let's consider the ground truth labels \(\boldsymbol{G} \in \mathbb{R}^{\mathrm{N}}\). The initial step involves using the association matrix $\boldsymbol{A}$ to filter the labels of all pixels under each centroid, computed as follows:
\begin{equation}
    \boldsymbol{G}_\text{f} = \boldsymbol{G}_\text{expand} * \boldsymbol{A}.
\end{equation}
Here, \(\boldsymbol{G}_\text{f} \in \mathbb{R}^{\mathrm{N} \times \mathrm{M}}\) denotes the filtered ground truth, \(\boldsymbol{G}_\text{expand}\) represents an expansion of \(\boldsymbol{G}\) across an additional dimension, and \(\ast\) signifies element-wise multiplication. Subsequently, we ascertain the occurrences of each class in the first dimension of \(\boldsymbol{G}_\text{f}\), yielding \(\boldsymbol{L} \in \mathbb{R}^{\mathrm{M} \times \mathrm{C}^\prime}\), with \(\mathrm{C}^\prime\) representing the number of distinct land cover classes. The pseudo-code for generating such labels is elaborated in the supplementary material.

\subsection{Learning Objective}
For the classification supervision of each spectral supertoken, we employ a cross-entropy loss function defined as:
\begin{equation}
    \mathcal{L}_\text{CE}(\hat{\boldsymbol{S}}, \boldsymbol{L})=-\frac{1}{\mathrm{M}}\sum_{m=1}^\mathrm{M}\sum_{c=1}^{\mathrm{C}'}\hat{\boldsymbol{S}}(m,c) \log \boldsymbol{L}(m,c),
\end{equation}
where \(\hat{\boldsymbol{S}}(m,c)\) represents the predicted probability of classifying the \(m\)-th token into the \(c\)-th category, and \(\boldsymbol{L}(m,c)\) corresponds to the respective soft label.

\section{Experiments}
We performed comprehensive comparative and ablation studies on hyperspectral image classification datasets. To assess our model's generalizability to downstream tasks, we also conducted extension experiments on the HS-SOD~\cite{HS-SOD} dataset for hyperspectral salient object detection. Results for HS-SOD are provided in the supplementary material.

\subsection{Implementation Details}
We optimize network parameters using the Adam optimizer~\cite{adam_optimizer} with a cosine annealing schedule~\cite{cos_annealing} for learning rate regulation. Experiments are conducted using PyTorch on an NVIDIA RTX 3090 GPU and Intel XEON Gold 5218R CPU. 
We segment semantic features, HSI, and spectral derivative features into $\mathrm{F} \times \mathrm{F}$ patches, generating $\mathrm{M}$ centroids per segment. $\mathrm{F}$ and $\mathrm{M}$ are set as 16 and 4, respectively. This results in 1024 clusters per image or feature. 
We evaluate three backbone networks in the Spatial-preserved Feature Encoder: ResNet18~\cite{ResNet} (DSTC-R), PVTv2-b1~\cite{PVT} (DSTC-P), and Swin-tiny~\cite{Swin} (DSTC-S).

\begin{table}[tp] \scriptsize
    \centering
    \caption{
    DSTC's F1-scores on Some Categories and Class Average F1-score Compared to Other Methods. `-R': ResNet18~\cite{ResNet}, `-P': PVTv2-b1~\cite{PVT}, `-S': Swin-tiny~\cite{Swin}.}
    \label{tab: quantitative}
    \setlength{\tabcolsep}{3.6pt}{}
    \begin{tabular}{l|ccccccccc|c}
    \toprule[1.2pt]
        Class & C1 & C2 & C4 & C7 & C9 & C11 & C15 & C18 & C23 & Avg. \\ 
    \midrule
        A2S2K~\cite{A2S2K} & 0.798 & 0.786 & 0.481 & 0.662 & 0.567 & 0.918 & 0.814 & 0.841 & 0.777 & 0.629 \\ 
        Capsule~\cite{Capsule} & 0.679 & 0.670 & 0.432 & 0.459 & 0.566 & 0.852 & 0.789 & 0.794 & 0.703 & 0.638 \\ 
        3D-CNN~\cite{3D-CNN} & 0.701 & 0.680 & 0.494 & 0.430 & 0.574 & 0.894 & 0.784 & 0.803 & 0.728 & 0.644 \\ 
        CLSJE~\cite{CLSJE} & 0.843 & 0.844 & 0.586 & 0.673 & 0.605 & 0.938 & 0.879 & 0.835 & 0.833 & 0.644 \\ 
        FreeNet~\cite{FreeNet} & 0.847 & 0.843  & 0.579 & 0.701 & 0.606 & 0.950 & 0.871 & 0.881 & 0.840 & 0.667 \\ 
        ViT~\cite{ViT} & 0.825 & 0.835 & 0.790  & 0.705 & 0.615 & 0.961 & 0.850 & 0.844 & 0.866 & 0.673 \\ 
        3D-FCN~\cite{3D-FCN} & 0.713 & 0.679 & 0.551 & 0.561 & 0.599 & 0.893 & 0.792 & 0.831 & 0.779 & 0.683 \\ 
    \midrule
        \rowcolor{Gray}
        DSTC-S~(Ours) & 0.853 & 0.848 & 0.595 & \textbf{0.730} & 0.617 & \textbf{0.970} & \textbf{0.880} & 0.884 & 0.856 & 0.718 \\ 
        \rowcolor{Gray}
        DSTC-R~(Ours) & 0.854 & 0.854 & 0.587 & 0.714 & 0.644 & 0.962 & \textbf{0.880} & 0.888 & 0.863 & 0.721 \\ 
        \rowcolor{Gray}
        DSTC-P~(Ours) & \textbf{0.858} & \textbf{0.859} & \textbf{0.600} & 0.714 & \textbf{0.653} & 0.944 & 0.876 & \textbf{0.894} & \textbf{0.885} & \textbf{0.723} \\  
    \bottomrule[1.2pt]
    \end{tabular}
\end{table}

\subsection{Results on WHU-OHS Dataset}
\nbf{Data and Experimental Setups}
The WHU-OHS dataset~\cite{WHU-OHS} consists of 7795 images acquired by the Orbita Hyperspectral Satellite (OHS) with 24 land cover classes. Each image in this dataset features a resolution of \(512 \times 512\) pixels and includes 32 spectral channels covering the 466-940~nm spectral range. For our training process, we configure the initial learning rate at \(5 \times 10^{-4}\), set the number of training epochs to 100, and establish a batch size of 8.

The performance of compared methods is evaluated using several metrics: class average F1-score (CF1), overall accuracy (OA), kappa coefficient (\(\kappa\)), and mean IOU (mIoU).

\nbf{Quantitative Results}
\cref{tab: quantitative} showcases the comparative results across various categories. In this context, `ViT' refers to an adapted version of the Vision Transformer~\cite{ViT}, wherein its classification head has been substituted with stacked convolutional layers to function as a decoder. These results underscore the superior classification performance of our DSTC. Notably, the DSTC-P variant attained a CF1 score of $0.723$, outperforming 3D-FCN by a margin of $0.040$. Furthermore, when integrating ResNet18 and Swin-tiny as backbones, DSTC's classification efficacy surpasses previously established methods. In terms of category-specific accuracy, DSTC exhibited outstanding performance, adeptly managing various land cover types. It is noteworthy that the majority of the compared methods are tailored for datasets with a relatively limited number of training samples. Consequently, some techniques do not exhibit optimal performance on the WHU-OHS dataset, which features a significantly larger volume of data.

\begin{table}[tp]
\centering \scriptsize
\caption{Quantitative Efficiency Analysis. `-R': ResNet18~\cite{ResNet}, `-P': PVTv2-b1~\cite{PVT}, `-S': Swin-tiny~\cite{Swin}.}
  \setlength{\tabcolsep}{13pt}{}
  \label{tab: efficiency analysis}
  \begin{tabular}{l | c | c c c }
    \toprule[1.2pt]
    Methods & CF1 & FLOPs (G) & \#Params (M) & Speed (FPS)\\
    \midrule
    A2S2K~\cite{A2S2K}  & 0.629 & 84.26  & \textbf{0.108} & 25.79 \\   
    CLSJE~\cite{CLSJE} & 0.644 & 412.20 & 5.352  & 6.51  \\
    3D-CNN~\cite{3D-CNN} & 0.644  & 1814.51 & 0.346 & \textbf{1122.88} \\
    FreeNet~\cite{FreeNet} & 0.667 & 116.20 & 2.508 & 231.17 \\
    \midrule
    \rowcolor{Gray}
    DSTC-S (Ours) & 0.718 & 19.78 & 11.589 & 9.61 \\
    \rowcolor{Gray}
    DSTC-R (Ours) & 0.721 & 17.87 & 4.001 & 10.52 \\
    \rowcolor{Gray}
    DSTC-P (Ours) & \textbf{0.723} & \textbf{16.15} & 9.032 & 9.79 \\
    \bottomrule[1.2pt]
  \end{tabular}
\end{table}

\begin{figure}[tp]
    \centering
    \includegraphics[width=1\linewidth]{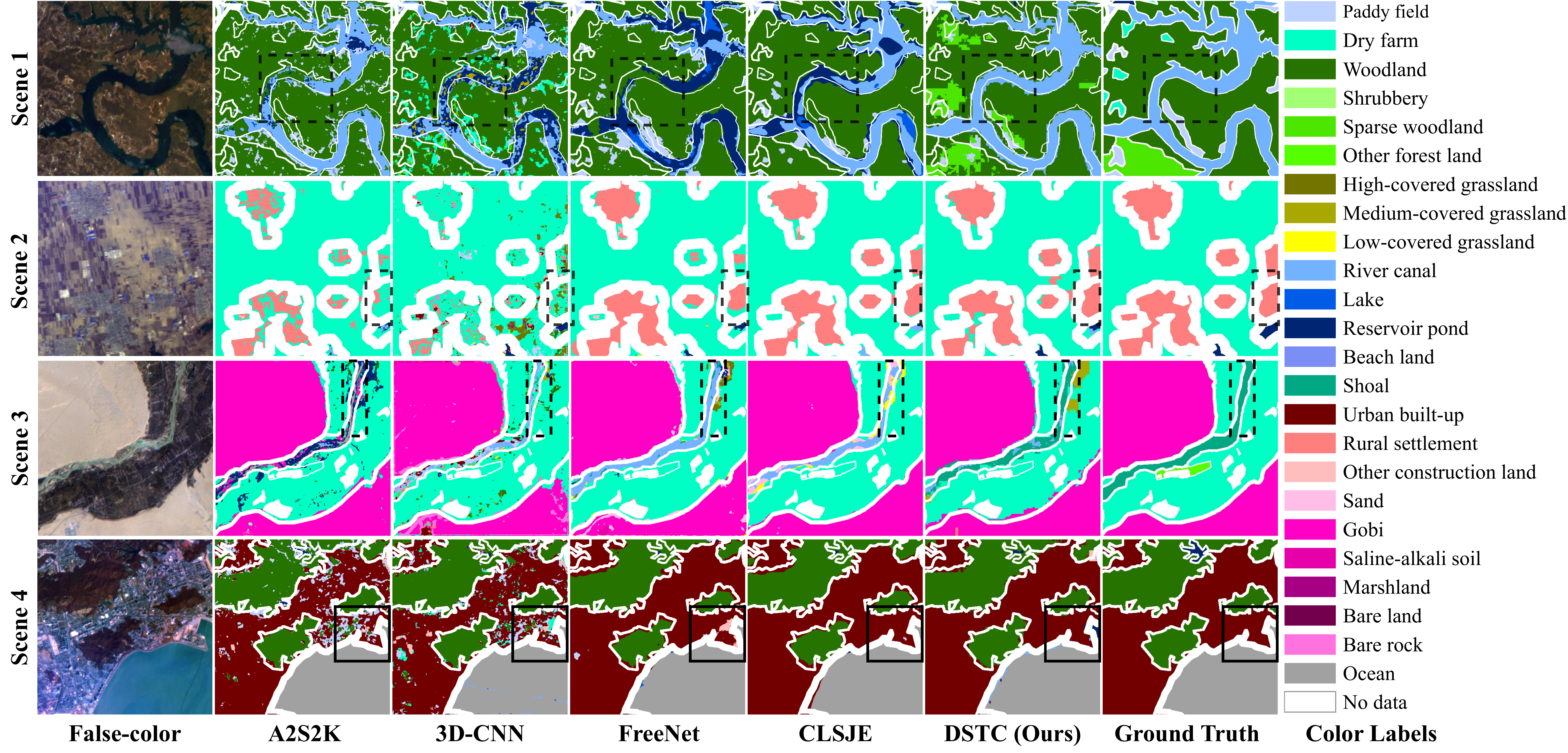}
    \caption{Qualitative result on WHU-OHS dataset. Our DSTC is closest to the ground truth, demonstrating the best classification capability.}
    \label{fig: result-vis}
\end{figure}

\nbf{Efficiency Analysis}
We have conducted a detailed efficiency analysis between our DSTC and other methods, focusing on Floating Point Operations (FLOPs), the number of parameters (\#Params), and the inference speed (FPS). The results, elaborated in \cref{tab: efficiency analysis}, indicate that DSTC significantly reduces FLOPs across various backbone configurations. For instance, DSTC-P requires $16.15$G FLOPs, which is only $3.92\%$ of the FLOPs required by CLSJE, and $0.89\%$ of those required by 3D-CNN. While DSTC's parameter count is not the lowest among the methods compared, the DSTC-R variant demonstrates a noteworthy reduction in parameters compared to CLSJE. Although the inference speed of DSTC may not be exceptionally high, it adequately fulfills the requirements for real-time inference. In conclusion, DSTC achieves real-time and accurate classification with minimal computational resource consumption.

\nbf{Qualitative Results}
\cref{fig: result-vis} presents the qualitative results from the WHU-OHS dataset, underscoring the superior classification capabilities of our DSTC, particularly in contiguous regions. This enhanced performance is notably evident in the precise delineation of the river bend, highlighted in black in the first scene, and the distinct representation of the rural settlement in the second scene. The effectiveness of DSTC can be largely ascribed to our pre-clustering technique, where regions with similar spectral and semantic features are consolidated into spectral supertokens. The classification predictions based on these tokens are crucial in preserving regional coherence, reducing pixel-level classification errors.

\begin{figure}[tp]
    \begin{minipage}[t]{0.48\textwidth}
        \centering
        \includegraphics[width=\linewidth]{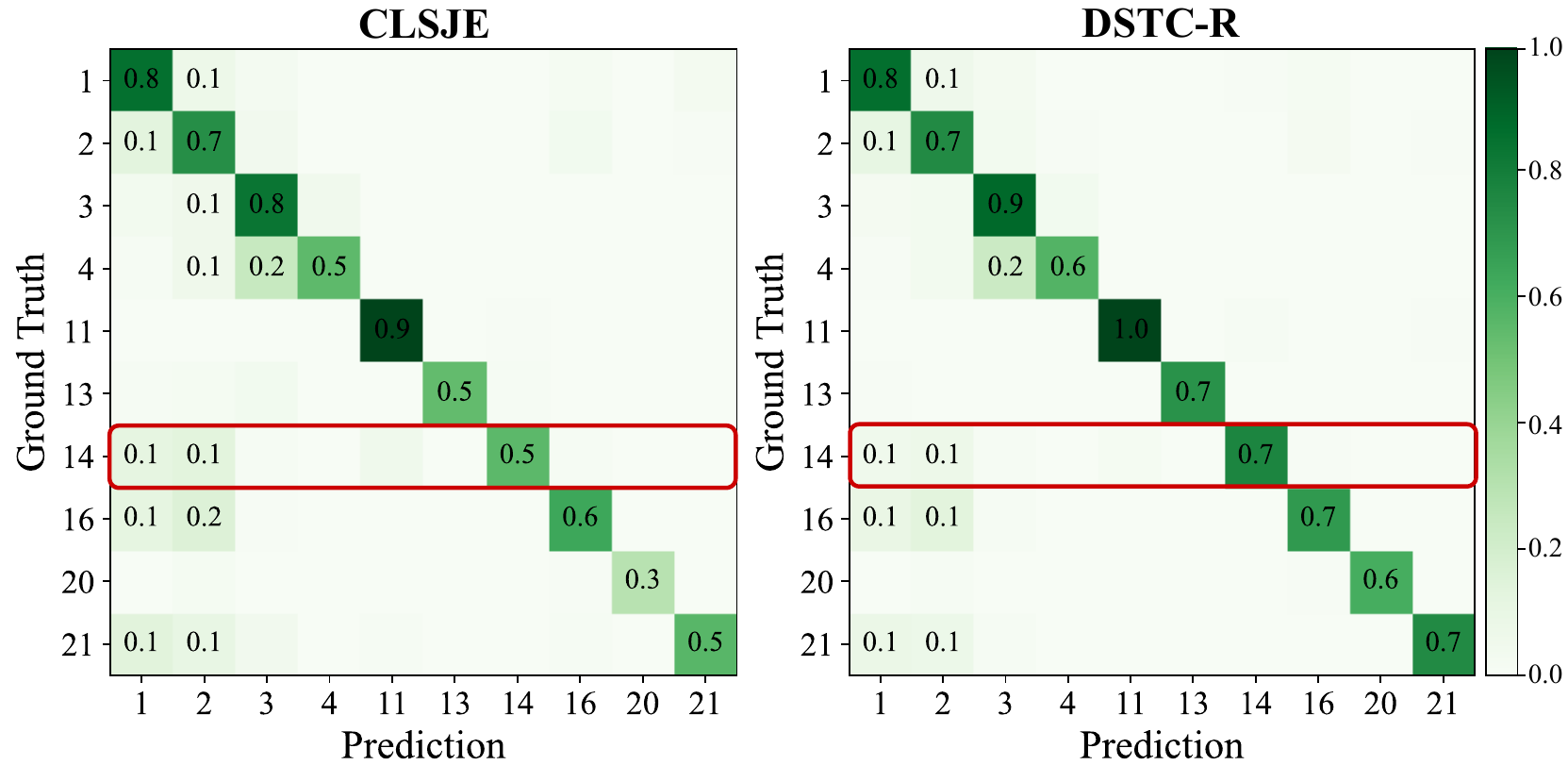}
        \figcaption{Confusion matrix of CLSJE and our DSTC-R on part class IDs.}
        \label{fig: confusion matrix}
    \end{minipage}
    \quad
    \begin{minipage}[t]{0.48\textwidth}
        \centering
        \includegraphics[width=\linewidth]{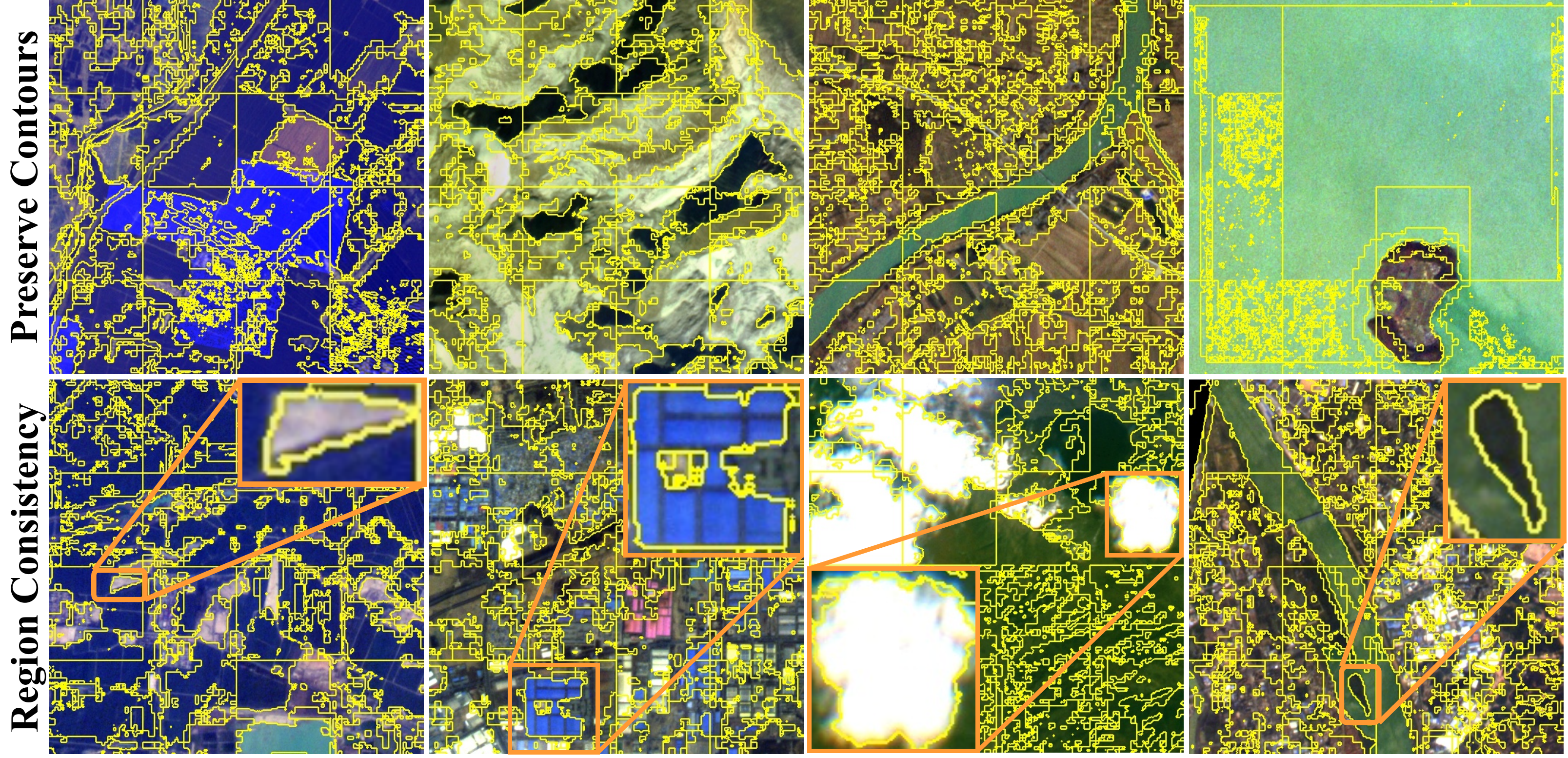}
        \caption{Visualization of clustered spectrally similar pixels. For better visualization, the crop factor is set to 4.}
        \label{fig: clustered pixels}
    \end{minipage}
\end{figure}

\nbf{Visualization of Confusion Matrix}
To emphasize the supremacy of our DSTC, we juxtapose its confusion matrix with that of CLSJE. The comparative results, displayed in~\cref{fig: confusion matrix}, exhibit darker diagonal elements in DSTC's matrix relative to CLSJE's. This signifies a higher accuracy in DSTC's classifications and a decrease in misclassifications. Notably, DSTC exhibits robust classification efficacy in the 14th category, "shoal", a result that aligns with the exceptional visualization outcomes observed in the third scenario depicted in~\cref{fig: result-vis}.

\nbf{Visualization of Clustered Similar Pixels}
\cref{fig: clustered pixels} clearly demonstrates the effectiveness of our spectrum-derivative-based pixel clustering algorithm in grouping spectrally similar pixels. The precise alignment of clustering boundaries with natural land cover delineations underscores our approach's accuracy in boundary detection. This enhanced boundary precision significantly contributes to improved classification accuracy.

\subsection{Experiment on IP, KSC, and UP Datasets}
\nbf{Data and Experimental Setups} 
The IP dataset contains $145 \times 145$ pixels, each with a spatial resolution of 20 $\mu m$. It includes 200 spectral bands and covers 16 distinct land cover categories, each comprising a broad range of sample sizes. The UP dataset has dimensions of $610 \times 340$ pixels and includes 115 spectral bands within the $0.43-0.86$ $\mu m$ wavelength range. It has a spatial resolution of $1.3$ $\rm m$ per pixel. After eliminating bands containing noise, 103 bands remain for classification. This dataset categorizes nine urban land cover types. The KSC dataset, with a spatial extent of $512 \times 614$ pixels, contains 176 spectral bands. We removed bands with low signal-to-noise ratios within the $400-2500$ $\rm nm$ wavelength range, resulting in a final count of 5202 samples across 13 categories.

For fair comparison, images are divided into overlapping $9 \times 9$ patches, with $10\%$ used for training and $90\%$ for testing. We align experimental settings with those in Li~\etal~\cite{CVSSN} for comparable patch-based HSI classification methods. Our DSTC uses a batch size of 16 and an initial learning rate of $1 \times 10^{-3}$. While comparison methods typically classify the central pixel of each patch, DSTC processes patches and outputs results matching input dimensions. Final pixel classification is determined by the most frequent result among all patches containing that pixel.

The efficacy is assessed through multiple metrics, including overall accuracy (OA), average accuracy (AA), and the kappa coefficient ($\kappa$).

\begin{table}[tp] \scriptsize
    \centering
    \setlength{\tabcolsep}{3.5pt}
    \caption{Quantitative results on IP, KSC, and UP datasets.}
    \label{tab: three benchmarks}
    \begin{tabular}{l| c  c  c | c  c  c| c  c  c }
    \toprule[1.2pt]
    \multirow{2}{*}{Methods} & \multicolumn{3}{c|}{IP} & \multicolumn{3}{c|}{KSC} & \multicolumn{3}{c}{UP} \\
    \cmidrule{2-10}
    &OA $\uparrow$ &AA $\uparrow$ &$\kappa$ $\uparrow$ &OA $\uparrow$ &AA $\uparrow$ &$\kappa$ $\uparrow$ &OA $\uparrow$ &AA $\uparrow$ &$\kappa$ $\uparrow$\\
    \midrule
    RSSAN~\cite{RSSAN} & 0.8300 & 0.8488 & 0.8053 & 0.9210 & 0.8665 & 0.9128 & 0.9913 & 0.9884 & 0.9885 \\
    SSAtt~\cite{SSAtt} & 0.9657 & 0.9654 & 0.9609 & 0.9681 & 0.9483 & 0.9645 & 0.9973 & 0.9969 & 0.9964 \\
    SSSAN~\cite{SSSAN} & 0.9511 & 0.9262 & 0.9442 & 0.0657 & 0.9447 & 0.9607 & 0.9983 & \textbf{0.9981} & 0.9978 \\
    SSTN~\cite{SSTN} & 0.9478 & 0.9230 & 0.9403 & 0.9698 & 0.9486 & 0.9664 & 0.9916 & 0.9876 & 0.9889 \\
    CVSSN~\cite{CVSSN} & 0.9827 & \textbf{0.9767} & 0.9802 & \textbf{0.9870} & 0.9787 & \textbf{0.9856} & \textbf{0.9988} & 0.9980 & \textbf{0.9983}\\
    \midrule
    \rowcolor{Gray}
    DSTC (Ours) & \textbf{0.9875} & 0.9615 & \textbf{0.9857} & 0.9860 & \textbf{0.9801} & 0.9844 & 0.9982 & 0.9969 & 0.9976 \\  
    \bottomrule[1.2pt]
    \end{tabular}
\end{table}

\begin{figure}[tp]
    \centering
    \includegraphics[width=1\textwidth]{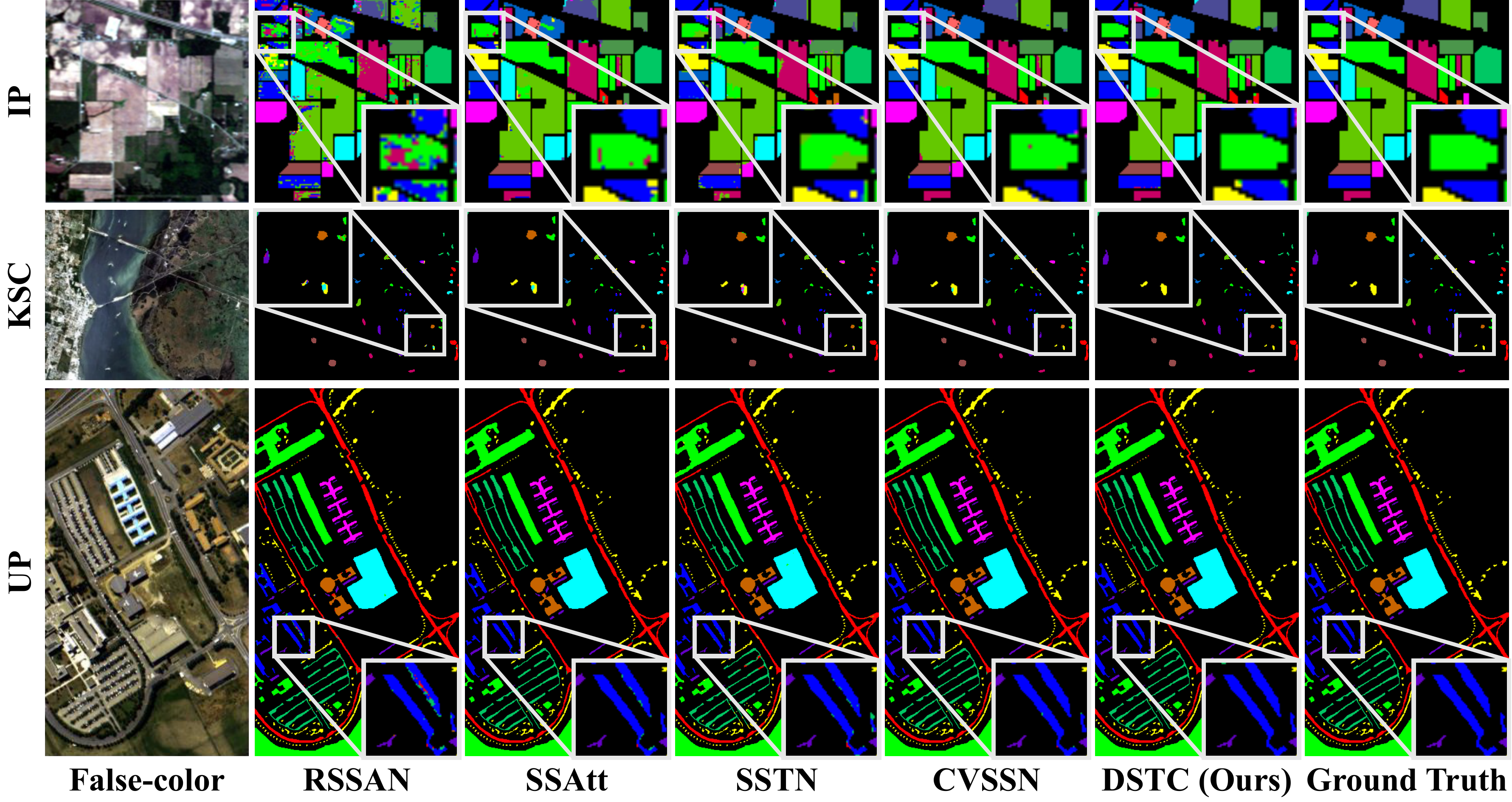}
    \caption{Qualitative results on IP, KSC, and UP datasets. Our DSTC pre-classifies pixels and ensures consistency within regions.}
    \label{fig: three-datasets}
\end{figure}

\nbf{Quantitative Results}
The numerical results from experiments conducted on the IP, KSC, and UP datasets are presented in~\cref{tab: three benchmarks}. These results validate that our DSTC delivers classification performance on par with CVSSN across multiple evaluation metrics on all three datasets, outperforming other comparative methods. Specifically, on the IP dataset, DSTC's OA and \(\kappa\) exceed CVSSN by $0.48\%$ and $0.55\%$ respectively, while the AA is $1.25\%$ lower than CVSSN. On the KSC dataset, DSTC's OA and \(\kappa\) are $0.1\%$ and $0.12\%$ lower than CVSSN respectively, with an AA $0.14\%$ higher. On the UP dataset, DSTC scores $0.06\%$, $0.11\%$, and $0.08\%$ lower across the three metrics respectively. Notably, our DSTC primarily employs spectral supertokens to pre-cluster the input HSI, targeting images with higher spatial resolution and datasets with larger volumn. Therefore, its performance on these three datasets is not as optimal as anticipated due to the less pronounced effect on the divided patches.

\nbf{Qualitative Results}
The classification results on the IP, KSC, and UP datasets are presented in~\cref{fig: three-datasets}, which include zoomed-in images for a more detailed view. Our DSTC exhibits robust classification performance across datasets with diverse characteristics, ranging from the geometrically regular IP dataset to the spatially disjoint KSC dataset. By leveraging spectrum-derivative-based pixel clustering, our DSTC groups spectrally similar pixel features into a single cluster. This approach ensures uniformity in classification results within contiguous areas, as demonstrated in the zoomed-in images.

\subsection{Ablation Study}
Ablation studies are conducted on the WHU-OHS dataset to validate the effectiveness of each component, with ResNet18~\cite{ResNet} serves as the backbone network.

\nbf{Hyperparameter Analysis}
We examine the influence of centroid numbers $\mathrm{M}$ and crop factor \(\mathrm{F}\) on our model's classification efficacy. The numeral results are listed in \cref{tab: centroid numbers} and \cref{tab: crop factor}, respectively. By analyzing the results under four evaluation metrics, we observe that the model's classification performance initially increases and then decreases with the augmentation of \(\mathrm{M}\) and \(\mathrm{F}\), peaking at values of $4$ and $16$, respectively.

\nbf{Effect of Semantic Feature}
By comparing the first two rows in \cref{tab: input data}, it's evident that semantic information (SF) significantly boosts the model's classification efficacy. This enhancement stems from incorporating semantic information into spectral supertokens creation. Such integration permits a nuanced similarity assessment between pixels and centroids. Consequently, it augments the representational power of spectral supertokens for adjacent pixels, thus elevating classification precision. 

\nbf{Effect of Spectral Derivative Features}
In \cref{tab: input data}, incorporating first-order derivatives (SD1) positively impacts the model's classification due to SD1's enhanced sensitivity to spectral shifts, which complements the original spectral data. However, second-order derivatives (SD2) negatively affect classification performance by introducing redundant information, as SD1 already provides sufficient discriminative features. Employing only SD2 results in slightly lower outcomes compared to only using SD1, with decreases of $0.001$ in both CF1 and $\kappa$, and a $0.002$ reduction in mIoU. Therefore, our approach excludes second-order spectral derivatives.

\begin{table}[tp]
    \begin{minipage}[t]{0.48\textwidth}
    \centering \scriptsize
        \tabcaption{Impact of Centroid Numbers $\mathrm{M}$.}
        \setlength{\tabcolsep}{7pt}{}
        \begin{tabular}{c|cccc} 
        \toprule[1.2pt]
        $\mathrm{M}$ & CF1 $\uparrow$ & OA $\uparrow$ & $\kappa$ $\uparrow$ & mIoU $\uparrow$ \\
        \midrule
        1 & 0.697 & 0.786 & 0.759 & 0.565 \\
        \rowcolor{Gray}
        4 & \textbf{0.721} & 0.803 & \textbf{0.779} & \textbf{0.591} \\
        9 & 0.708 & \textbf{0.804} & \textbf{0.779} & 0.580 \\
        16 & 0.685 & 0.792 & 0.765 & 0.552 \\
        \bottomrule[1.2pt]
        \end{tabular}
        \label{tab: centroid numbers}
    \end{minipage}
    \quad
    \begin{minipage}[t]{0.48\textwidth}
        \centering \scriptsize
        \tabcaption{Impact of Crop Factor $\mathrm{F}$.}
        \setlength{\tabcolsep}{7pt}{}
        \begin{tabular}{c|cccc}
        \toprule[1.2pt]
        $\mathrm{F}$ & CF1 $\uparrow$ & OA $\uparrow$ & $\kappa$ $\uparrow$ & mIoU $\uparrow$ \\
        \midrule
        4 & 0.698 & 0.773 & 0.743 & 0.564 \\
        8 & 0.719 & 0.797 & 0.771 & 0.589 \\
        \rowcolor{Gray}
        16 & \textbf{0.721} & \textbf{0.803} & \textbf{0.779} & \textbf{0.591} \\
        32 & 0.711 & 0.801 & 0.776 & 0.581 \\
        \bottomrule[1.2pt]
        \end{tabular}
        \label{tab: crop factor}
    \end{minipage}
\end{table}

\begin{table}[tp]
    \begin{minipage}[t]{0.48\textwidth}
    \centering \scriptsize
    \tabcaption{Effect of Semantic Feature and Orders of Spectral Derivatives.}
    \label{tab: input data}
    \setlength{\tabcolsep}{3pt}{}
    \begin{tabular}{ccc|cccc}
    \toprule[1.2pt]
        SF & SD1 & SD2 & CF1 $\uparrow$ & OA $\uparrow$ & $\kappa$ $\uparrow$ & mIoU $\uparrow$ \\
    \midrule
        \nohave & \nohave & \nohave & 0.669  & 0.761 & 0.730 & 0.535 \\ 
        \have & \nohave & \nohave & 0.717  & \textbf{0.804} & \textbf{0.779} & 0.587 \\ 
        \have & \have & \have & 0.713 & 0.801 & 0.776 & 0.584 \\
        \have & \nohave & \have & 0.720 & 0.803 & 0.778 & 0.589 \\
        \have & \have & \nohave & \textbf{0.721} & 0.803 & \textbf{0.779} & \textbf{0.591} \\  
    \bottomrule[1.2pt]
    \end{tabular}
    \end{minipage}
    \quad
    \begin{minipage}[t]{0.48\textwidth}
    \renewcommand\arraystretch{1.5}
    \centering \scriptsize
    \tabcaption{Impact of Class-proportion-based Soft Labels.}
    \label{tab: CPSL}
    \setlength{\tabcolsep}{4pt}{}
    \begin{tabular}{l|cccc}
    \toprule[1.2pt]
        Supervision & CF1 $\uparrow$ & OA $\uparrow$ & $\kappa$ $\uparrow$ & mIoU $\uparrow$ \\
    \midrule
        Hard Labels & 0.443  & 0.485 & 0.406 & 0.298 \\ 
        Dense-CE & 0.699  & 0.789 & 0.763 & 0.567 \\ 
        Soft Labels & \textbf{0.721} & \textbf{0.803} & \textbf{0.779}  & \textbf{0.591} \\
    \bottomrule[1.2pt]
    \end{tabular}
    \end{minipage}
\end{table}

\nbf{Impact of Class-proportion-based Soft Labels}
To investigate the effectiveness of class-proportion-based soft labels, we conducted an experiment replacing soft labels with hard labels, assigning a single category to each spectral supertoken. We also reverted these supertokens to their original image states and used Cross-Entropy loss for pixel-level supervision, denoted as dense-CE. As detailed in \cref{tab: CPSL}, supervising spectral supertokens with hard labels reduces classification performance, likely due to exacerbating data distribution imbalances inherent in single-category labels. In contrast, using soft labels significantly improved performance across all four metrics compared to dense-CE, highlighting the effectiveness of class-proportion-based soft labels.

\section{Conclusion}
In this work, we introduce the innovative Dual-stage Spectral Supertoken Classifier (DSTC) designed for hyperspectral image classification. At the heart of our approach is the spectrum-derivative-based pixel clustering algorithm, which adeptly groups spectrally similar pixels into spectral supertokens. This pre-classification step plays a crucial role in enhancing the model's resilience to minor spectral variations and in achieving precise boundaries within contiguous areas. Additionally, we introduce a novel class-proportion-based soft label to supervise each spectral supertoken. Experimental results on WHU-OHS, IP, KSC, and UP datasets highlight DSTC's superiority over previous methods, demonstrating its effectiveness. Further experiments conducted on a hyperspectral salient object detection dataset, HS-SOD, validate the generalizability of DSTC.

\section*{Acknowledgments}
This work was financially supported by the National Key Scientific Instrument and Equipment Development Project of China (No. 61527802), the National Natural Science Foundation of China (No. 62101032), the Young Elite Scientist Sponsorship Program of China Association for Science and Technology (No. YESS20220448), and the Young Elite Scientist Sponsorship Program of Beijing Association for Science and Technology (No. BYESS2022167).

%
%
\bibliographystyle{splncs04}
\bibliography{egbib}
\end{document}